\documentclass{INTERSPEECH2023}
\usepackage{tikz}
\usepackage{graphicx}
\usepackage{subfig}
\usepackage{floatrow}
\usepackage{multirow}

\usepackage{multirow}
\usepackage{svg}
\usetikzlibrary{arrows,positioning,arrows.meta, calc}
\usepackage{ifthen}
\definecolor{darkgreen}{RGB}{11, 153, 56}

\definecolor{extradarkgreen}{RGB}{22, 199, 78}
\usepackage{dirtytalk}

\interspeechcameraready 

\title{Leveraging Cross-Utterance Context For ASR Decoding}
\name{Robert Flynn, Anton Ragni}

\address{Department of Computer Science, University of Sheffield, 211 Portobello, Sheffield S1 4DP, UK}
\email{rflynn2@sheffield.ac.uk, a.ragni@sheffield.ac.uk}

\begin{document}

\maketitle
 
\begin{abstract}
While external language models (LMs) are often incorporated into the decoding stage of automated speech recognition systems, these models usually operate with limited context. Cross utterance information has been shown to be beneficial during second pass re-scoring, however this limits the hypothesis space based on the local information available to the first pass LM. In this work, we investigate the incorporation of long-context transformer LMs for cross-utterance decoding of acoustic models via beam search, and compare against results from n-best rescoring. 
Results demonstrate that beam search allows for an improved use of cross-utterance context. When evaluating on the long-format dataset AMI, results show a 0.7\% and 0.3\% absolute reduction on dev and test sets compared to the single-utterance setting, with improvements when including up to 500 tokens of prior context. Evaluations are also provided for Tedlium-1 with less significant improvements of around 0.1\% absolute.

\end{abstract}

\vspace{-2mm}
\section{Introduction}

The decoding stage of end-to-end automated speech recognition (ASR) systems often benefits from the use of an external language model. For Connectionist temporal classification (CTC) \cite{graves2006connectionist} based acoustic models (AM) this is often crucial for good performance due to their conditional independence assumption. Traditionally, $n$-gram models have been used for this task, although due to data scarcity these models are usually restricted to a limited context of 3-4 preceding words. Consequently, neural models are often applied during second pass \cite{sun2021transformercrossutt, irie2019languagerwthdeeptlm, huang2019empirical, parthasarathy2019longspanlm, xiong2018microsoft2017, xiong-etal-2018-session, sun2020crosserrorsample, chiu2021crossuttbert} or first pass decoding \cite{chen2020lstmlmfirstpass, toshniwal2018comparison, zenkel2017comparison, hwang2016character, baevski2020wav2vec} stages for further reductions in word error rate (WER).

The advent of the transformer architecture \cite{vaswani2017attention} has lead increased performance on many tasks, including ASR. In part, this can be attributed to its ability to effectively propagate gradients across long-distances, enabling the learning of much longer-term dependencies. The use of external transformer language models (TLMs) to improve ASR has been investigated in prior work \cite{irie2019languagerwthdeeptlm, pandey2022lattention, sun2021transformercrossutt, huang2019empirical}, with results demonstrating an advantage over recurrent-based architectures. 

While many realistic use cases for ASR will involve long and continuous conversations or talks, ASR systems are mainly trained over very short utterances with a (often invalid) $i.i.d$ assumption. Likewise decoding/re-scoring is typically performed over individual utterances, which can lead to context fragmentation  reductions in performance.

The contributions made as part of this work are as follows: 1. We assess the benefit of cross-utterance information for decoding, including how much context is useful and the impact of errors in the history. 2. We demonstrate that external language model integration via beam search improves the ability to utilise cross-utterance information compared to re-scoring. 3. A set of adaptions from prior work \cite{multiquerynoam, dai-etal-2019-transformerxl} are proposed for use in this task making shallow fusion with TLMs more feasible.

The remainder of the paper is organized in the following manner: Section \ref{sec:related_work} provides a brief overview of related work, Section \ref{sec:method} details our method including adaptions made for efficient decoding (\ref{sec:effdec}) and language modelling in a conversational setting (\ref{sec:conversationLM}). Experimental details such as model and decoding configuration, and datasets is provided in section \ref{sec:experimental}. Results and key findings are given in Section \ref{sec:results} with our conclusion in Section \ref{sec:theend}.

\vspace{-2mm}
\section{Related work}
\label{sec:related_work}
Long-range linguistic context has been previously been exploited in ASR through the use of external neural language models. In \cite{sun2021transformercrossutt, parthasarathy2019longspanlm, sun2020crosserrorsample, chiu2021crossuttbert} cross-utterance information is found to be beneficial for perplexity (PPL) and WER during second-pass re-scoring with LSTM and transformer type models. Similar findings are presented in \cite{xiong-etal-2018-session} where LSTMs are adapted for a conversational setting and effectively used for re-scoring with an extended history. However it is not clear from these works how much context is useful for re-scoring. First pass cross-utterance decoding using LSTM models has previously been explored \cite{chen2020lstmlmfirstpass} using on-the-fly composition with a WSFT \cite{mohri2002weightedwsft} based decoder. This work shows some performance benefit for using the LSTM during the initial pass compared to re-scoring. No benefit is found to using greater than 4 sentences of prior context, which may be due limitations of LSTMs use of context \cite{khandelwal2018sharp}.

\vspace{-2mm}
\section{Transformer Decoding}
\label{sec:method}
\subsection{Transformer Language Modelling}
While typically a transformer \cite{vaswani2017attention} may consist of a bi-directional encoder, followed by a causal decoder, for language modelling we use the decoder-only variant transformer. This consists of alternating multi-headed self-attention with a causal mask, and feed-forward modules. 

Given a word sequence $\boldsymbol{w} = (w_1 ,..,w_{T})$ causal language models are trained to estimate the conditional probability of $P(w_t | \boldsymbol{w}_{<t})$. Word sequence probabilities can then be obtained by an expansion resulting in $P(\boldsymbol{w})$, when working with log-likelihoods this equates to: $\sum_{t=1}^{T}\log P(w_t|\boldsymbol{w}_{<t})$. For the purpose of decoding these likelihoods can be treated as scores and combined with the AM through a log-linear interpolation. 

\vspace{-2mm}
\subsection{Self-Attention}
Self attention, which is one of the key components of TLMs, is a \textit{mixing} operation between input tokens that enables the learning of local and long-range statistical patterns. The input $X$ is first transformed to obtain the query, key and value matrices. 
\begin{equation}
    Q, K, V = XW^Q, XW^K, XW^V
\end{equation}
In this work we use the key query normalisation variant of attention \cite{henry2020keyquerynorm}.
\begin{equation}
    \hat{q}_i = \frac{q_i}{\lVert q_i \rVert}, \hat{k}_i = \frac{k_i}{\lVert k_i \rVert}
\end{equation}
This involves applying $L_2$ normalisation to the keys and queries, obtaining their dot product, and scaling the result by a learnt parameter $g$.
To encode positional information information an additional matrix $P$ is added as a bias to the similarity matrix, alongside a causal mask $M$ to prevent interactions with future tokens, before the softmax is applied. The resultant distribution is used to obtain a weighting of the value vectors, which is used to produce inputs to the following layer.
\begin{equation}
   \mathrm{Attention}(\hat{Q}, \hat{K}, V) =   \mathrm{Softmax}(g \cdot \hat{Q}\hat{K}^{T}+P+M)V
\end{equation}

\subsection{Adaptions for Efficient Decoding}
\label{sec:effdec}
\subsubsection{Key-Value Caching}
Unlike recurrent-based networks TLMs do not condense the history into a single representation, and instead process their input in parallel. For each token or utterance passed to the model we want to avoid re-computing the previous history as this would be extremely costly. As the language model is causal, and prior states are not updated based on future information, it is possible to cache keys and values from the attention sub-layers to avoid re-computation. This is not an approximation, and can be viewed as computing the attention in a sequential manner which is explored in prior work \cite{dai-etal-2019-transformerxl, press2020shortformer}. This enables efficient cross-utterance decoding using TLMs with a linear increase in memory for each token in the cache.

To achieve this we simply concatenate keys $\hat{K}$ and values $V$ from previous timesteps $t_{0:t-1}$ that are cached during each forward pass with the data from the current timestep.
\begin{equation}
    \hat{K}_{0:t}, V_{0:t} = (\hat{K}_{0:t-1}, \hat{K}_t), (V_{0:t-1}, V_t)
\end{equation}
With $\hat{Q}_t, \hat{K}_{0:t}, V_{0:t}$ acting as input to equation 3 for this method.
To limit the sequence length at inference time, the cache can be truncated, when doing this we find it necessary to maintain the beginning of sentence token at the start of the history.

\vspace{-2mm}
\subsubsection{Multi-Query Attention}
Additionally, we employ \textit{Multi-Query} Attention \cite{multiquerynoam}, which uses only 1 head for both the keys and the values. This reduces slowdown due to memory bandwidth, with negligible performance degradation, and was recently validated in large-scale training of PaLM \cite{chowdhery2022palm}. When decoding incrementally with a batch size of 25 and cache/history size of 500 tokens, this improves our iteration time by \textbf{74\%}.

\vspace{-2mm}
\subsection{Conversational Language Modelling}
\label{sec:conversationLM}
\begin{figure}[!h]
\includegraphics[width=8cm]{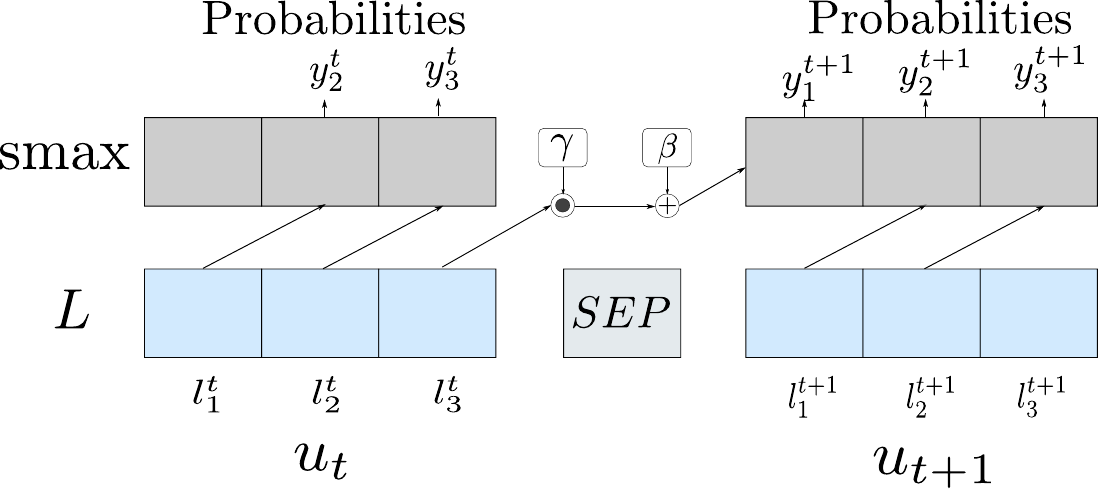}
\caption{Modifications for language modelling in a conversational setting.}
\label{fig:convlming}
\end{figure}

As the language model is operating over series of utterances $U = (u_0, ... u_t)$, where utterance boundaries may indicate speaker changes, or changes in topic we choose to model this with a few modifications to the architecture. These modifications are illustrated in figure \ref{fig:convlming}.

A separator token $SEP$ is introduced to denote utterance boundaries, this is passed to the model alongside the input at the end of each utterance. There are no targets for this token and it is simply used to denote a boundary in the history.

To predict the first token in a following utterance, we introduce initial token prediction, where learnt per dimension scalar $\gamma$ and offset $\beta$ transformations are applied to the logit predictions from the last token in the current utterance. Specifically, given a sequence of logits $L^t  = (l^t_1 ... l^t_n)$ from utterance $u_t$, probabilities for the first token in the following utterance $y_0^{t+1}$ are obtained as follows:

\begin{equation}
    y^{t+1}_1 = \mathrm{Softmax}(l^t_n \odot \gamma + \beta)
\label{eq:crossutt}
\end{equation}
Hence the model can learn to modulate probabilities for tokens that may be more, or less likely at the beginning of an utterance. The inclusion of eq. \ref{eq:crossutt} was motivated by the frequent speaker changes in AMI \cite{kraaij2005ami}, with new utterances often beginning with a hesitation such as \say{hmm}.  

\vspace{-2mm}
\subsection{Beam Search}
Language models can be combined with the output probabilities of an AM using beam search. Ideally, we would like to find the most probable path across all alignments using all models. However, this is infeasible so we approximate by restricting the search to a set of top paths or beams.
For each time-step $t$ we use the AM probabilities and look-ahead probabilities from the TLM to score the vocabulary indices at that time-step $i_t$.
\begin{equation}
    \mathrm{Score}(i_t) = \log P_{AM}(i_t) +  \mathrm{Score}_{LM}(i_t)
\end{equation}
Our AM uses the CTC decoding algorithm \cite{graves2006connectionist} which features blank tokens $\theta$ as part of the vocabulary, and allows for repetitions which are collapsed as part of decoding. As the TLM does not provide probabilities for repeats $i_t = i_{t-1}$ or blanks $\theta$ an insertion bonus $\beta$ is used to prevent the search favouring these tokens. Additionally, a scaler $\alpha$ is used to weight the TLMs score. Hence the TLMs score is as follows \cite{zenkel2017comparison}:

\begin{equation}
    \mathrm{Score}_{LM}(i_t) =  \begin{cases}
      0, & \text{if $i_t=\theta$ or $i_t = i_{t-1}$} \\
      \alpha \log P_{LM}(i_t) + \beta, & \text{otherwise}
    \end{cases}
\end{equation}

\vspace{-2mm}
\section{Experimental Setup}
\label{sec:experimental}
\subsection{Model and Training Specifications}
The TLMs all have 12 layers with a hidden dimension of 256. For the feedforward blocks, SwiGLU layers \cite{shazeer2020glu} are used with an expansion factor of 4. Key query normalised attention is used in-place of dot product attention as in \cite{henry2020keyquerynorm}. 8 heads are used for the queries during attention with 1 head for the keys and values \cite{multiquerynoam}. For positional encoding, the \textit{dynamic position bias} proposed in \cite{wang2021crossformer, liu2022swin} is used. TLMs are trained without dropout, as they under-fit the pre-training corpus. In total the TLMs amount to around 11M parameters. 

During training of the TLM windows of up to 25 prior utterances are used as context. A TLM is pre-trained and fine-tuned for each dataset, in order to match the AMs vocabulary and training takes around 5 days on a GTX 3060 GPU.

For the AM, we use an encoder-only Conformer architecture \cite{gulati2020conformer} trained with self-conditioned CTC and intermediate losses \cite{nozaki2021relaxingselfconditioned}. Batch normalisation \cite{ioffe2015batch} is swapped out in favour of batch renormalisation \cite{ioffe2017batchrenorm}. Key query normalised attention and dynamic position biases are also used with the AM. For both AMs a byte pair vocabulary of size 128 is used.
For the Tedlium dataset, the model has 12 layers, a hidden dimension of 256, 8 attention heads, and a convolution kernel width of 31. For AMI a smaller hidden dimension of 176 is used with 4 attention heads, 16 layers, and a convolution kernel width of 15. SpecAugment \cite{park2019specaugment} is used during training alongside dropout ($0.1$ on feedforwards and $0.3$ post-attention) for regularisation. Training of the AMs takes around 2 days on AMI and 4 days on Tedlium using a V100 GPU.

Models are trained using the Madgrad optimizer \cite{defazio2022adaptivitymadgrad} with weight decay of $1e-6$. An EMA of model parameters set to 0.9999 is used for the AM, and during finetuning of the TLM.

\vspace{-2mm}
\subsection{Beam Search}
For beam search a beam width of 25 is used. The search is constrained at each time-step to indices within a given threshold of the argmax of the AM probabilities (cut-off threshold). The insertion bonus $\beta$, TLM weighting $\alpha$ and the cut-off threshold were trained through a random search. The ranges explored in the search are as follows: For $\alpha$ $[0.0,1.0]$, $\beta$ $[-0.1,0.8]$, and $[-4, -12]$ for the cutt-off threshold. Only the top beam is maintained when transferring across utterance boundaries, this helps reduce the propagation of errors found in low probability beams from previous utterances.
The beam search is parallelised over each utterance in the single utterance setting, or over each meeting/talk for cross-utterance evaluation using the ray library \cite{moritz2018ray}. Single-utterance decoding takes around 10 minutes and 40 minutes on Tedlium and AMI, and 35-55 minutes and 2.5-3.5 hours in the cross-utterance setting, however our implementation is not optimized and is implemented in python.

\vspace{-2mm}
\subsection{N-best Rescoring}
To obtain the N-best list, beam search with a width of 1000 is used with a n-gram trained with the same BPE vocabulary as the AM, this is then re-ranked through a word-level n-gram model and pruned to an n-best list of 100 hypotheses.
The TLM is combined with first-pass models through weighted addition of the log probabilities. Additionally, a length penalty is included. These weights/hyperparameters are trained through a random search. For best performance it was necessary to standardise the log probabilities of the TLM using mean and standard deviation statistics from the top hypotheses. 

\vspace{-2mm}
\subsection{Datasets}

\subsubsection{Tedlium 1}

Tedlium \cite{rousseau2012tedlium} is an ASR corpus consisting of single-speaker TED talks. This data is selected as it is an ideal use-case for long-range ASR, with many unique speakers using varied language, talking continuously for periods of up to 18 minutes. Despite this there is, too our knowledge, no work that attempts to fully utilise this longer time-frame for ASR. For this work we use the smaller $1^{st}$ release of Tedlium. For pre-processing, spaces between the apostrophe in contraction are removed. In this data there are instances of large un-annotated gaps between utterances (this is alot less prevalent in the test and dev data). To address this, when training cross-utterance TLMs the cache is not propagated over gaps larger than ten seconds. Statistics for this corpus are provided in table \ref{tab:ami_ted_stats}.

\begin{table}[hbt]
    \centering
    \begin{tabular}{c|c|c|c}
              & Total Words & Duration (h) & Recordings  \\\hline
            Train & 764,088 / 1,315,797  & 80.1 / 118.1 & 132 / 774 \\
            Dev &  95,374 / 17,733 & 9.8 / 1.6 & 18 / 8 \\
            Test & 89,978 / 27,500 & 9.4 / 2.6 & 16 / 11
    \end{tabular}
    \caption{Corpus statistics for AMI / Tedlium-1}
    \label{tab:ami_ted_stats}
\end{table}

\begin{table*}[h!]
\centering
\begin{tabular}{c|c|c|c|c|c|c}

Dataset & 0 & 50 & 100 & 250 & 500 & 1000 \\\hline
AMI & 89.77 / 75.49  & 74.37 / 67.82 & 71.43 / 66.12 & 68.18 / 64.47 & 66.53 / 63.80 & \textbf{64.89} / \textbf{62.54} \\
Tedlium & 148.72 / 136.21 & 132.47 / 114.60  & 126.79 / 118.96   & 119.70 / 108.10  & 117.86 / 104.63 & \textbf{116.33} / \textbf{101.91} \\

\end{tabular}

\caption{Perplexity (PPL) for TLMs on dev / splits at (0, 50, 100, 250, 500) tokens of context from preceding utterances}
\label{tab:ppls}
\end{table*}

\begin{table*}[h!]
\centering
\begin{tabular}{c|c|c|c|c|c|c}

Dataset & Decoding Method & 0 & 50 & 100 & 250 & 500 \\
\hline
\multirow{4}{*}{\rotatebox[origin=c]{0}{AMI}} 
& Rescoring & 20.96 / 19.41 & \textbf{20.88} / 19.31 & 20.88 / 19.31 & 20.89 / \textbf{19.30} &  20.88 / 19.30 \\
& Rescoring (GTH) & 20.96 / 19.41 & 20.84 / 19.26 & \textbf{20.83} / 19.26 & 20.83 / 19.23 &  20.83 / \textbf{19.21} \\
& Beam Search & 20.52 / 18.98 & 19.82 /  18.75 & \textbf{19.78} / 18.72  & 19.78 / 18.67 & 19.80 / \textbf{18.65} \\
& Beam Search (GTH) & 20.52 / 18.98 & 19.71 /  18.68 & 19.70 / 18.65  & 18.67 / 18.62 & \textbf{19.66} / \textbf{18.57} \\
\hline
\multirow{4}{*}{\rotatebox[origin=c]{0}{Tedlium}} 
& Rescoring & 9.51 / 8.48 & 9.52 / 8.44 & 9.52 / 8.45 & \textbf{9.50} / \textbf{8.44} & 9.52 / 8.45 \\
& Rescoring (GTH) & 9.51 / 8.48 & 9.48 / 8.42 & 9.47 / \textbf{8.41} & \textbf{9.44} / 8.42 & 9.50 / 8.42 \\
& Beam Search & 9.73 / 8.61 & 9.69 / 8.54 & 9.69 / 8.48 & \textbf{9.67} / \textbf{8.47}  & 9.72 / 8.48 \\
& Beam Search (GTH) & 9.73 / 8.61 & 9.62 / 8.52 & 9.59 / 8.48 & \textbf{9.56} / \textbf{8.35}  & 9.61 / 8.41 \\

\end{tabular}
\caption{Results (WER) for each decoding method on AMI and Tedlium with (0, 50, 100, 250, 500) tokens of context from preceding utterances. Results are given for dev / test sets. GTH denotes the use of the Ground Truth transcripts to form the History.}
\label{tab:mainresults}
\end{table*}
\begin{table}[h!]
    \centering 
    \begin{tabular}{l|c|c}
        Dataset & Dev  & Test \\\hline
        Tedlium & 11.73 / 9.89 &  10.37 / 8.80    \\
        AMI       &  22.94 / 21.09  &  21.49 / 19.58 \\
    \end{tabular}
    \caption{Baselines WERs for each dataset, results are given without/with an n-gram language model}
    \label{tab:fpbaselines}
\end{table}

\vspace{-3.5mm}
\subsubsection{AMI}

AMI \cite{kraaij2005ami} consists of multi-speaker meetings. The individual headphone microphone (IHM) version is used in this work. This data was selected due to it's long-format conversational setting and it's use in similar, prior work. However, features of this dataset such as, very frequent speaker changes and speaker overlaps, may present difficulties for measuring the long-range performance of ASR systems in isolation. 

Utterances in the training data longer than 16 seconds are re-segmented using the provided time-stamps, while dev and test splits are kept intact. Statistics are provided in table \ref{tab:ami_ted_stats} 
\vspace{-2mm}
\subsubsection{OpenSubtitles}
As a pre-training corpus for the TLMs OpenSubtitles\footnote{\url{http://www.opensubtitles.org/}} \cite{lison2016opensubtitles2016}, is used. We pretrain on a subset of the corpus for containing 408,985,555 words for one epoch. Numbers and monetary values are converted to their orthographic form, and all text is set to lower case and punctuation excluding apostrophes is removed. 

\vspace{-2mm}

\section{Experimental Results}
\label{sec:results}

Perplexities for the TLMs are provided in table \ref{tab:ppls}. Here the models show continued decreases in PPL which begins to plateau at 1000 tokens of prior context.

Table \ref{tab:fpbaselines} presents our baseline results for both greedy decoding, and the initial first-pass which is used in re-scoring. Notably, our baselines achieve good results, performing better on both datasets than other single-domain AMs in the literature which do not incorporate i-vectors \cite{sun2021transformercrossutt, chiu2021crossuttbert, yang2022onlinecontinual}.  

Results for re-scoring and beam search are given in table \ref{tab:mainresults}. The presentation of results for \textbf{AMI} is split into sub-sections that are based on the key components of this work, lastly we overview and discuss performance on \textbf{Tedlium} separately.

\vspace{-2mm}
\subsection{Re-scoring vs Beam Search}

For decoding via beam search on AMI we see sizeable improvement over re-scoring, both in the utilisation of context, and more generally in the single-utterance setting. When zero prior context is available we see a 2.4\% and 2.5\% (dev and test) absolute decrease over the greedy decoding baseline for beam search, compared to 2.0\% and 2.1\% for re-scoring.

With additional context we find much more consistent improvements with beam search. For example, with the full 500 tokens of prior context beam search shows an absolute decrease over the single-utterance setting of 0.7\% and 0.3\% for dev and test respectively. For re-scoring 0.1\% and 0.1\% absolute decreases are attained with the additional context. Hence we find that re-scoring is limiting the use of the context with improvements from beam search becoming more pronounced in the cross-utterance setting.

\vspace{-2mm}
\subsection{How much context is useful?}
Unsurprisingly, on the AMI corpus, decoding benefits most from including the recent history of the previous 50 tokens, with gains of 0.7\% and 0.2\% (dev and test) absolute over the single-utterance setting using beam search. On average there is 23.7 tokens per utterance on AMI, hence this amounts to around 2 prior utterances of context. For the test set we can see continued improvements with up to 500 tokens of prior context with a further 0.1\% absolute decrease when over 50 tokens of context. This is on average 21 utterances of context that we are able to benefit from, increasing beyond 500 tokens brought no further improvements. On the dev set while we see significant gains from the cross-utterance context of 0.7\% absolute there is no benefit from increasing the context beyond 100 tokens. 
\vspace{-2mm}
\subsection{Impact of errors in the context}
During the training the model is provided with ground truth transcripts, while during decoding, only the decoded history is available. To examine the effect of this mismatch on the use of the context we present additional results where during decoding the transcripts from previous utterances are used as the context. This is denoted as Ground Truth Histoy (GTH) in table \ref{tab:mainresults}.

For both dev and test dataset on AMI when decoding via beams search we see around an additional 0.1\% absolute decrease when using the GTH across all context lengths. Additionally, with the GTH we see continued improvements on the dev set until 500 tokens of context, while otherwise this plateaus at 100 tokens. Similar findings are seen for re-scoring. As such, we find that test-time errors have a consistent impact on the use of the history. This does not account for the impact of errors within the current utterance, which is likely more severe.

\vspace{-2mm}
\subsection{Performance on Tedlium}
On the Tedlium data re-scoring provides an improvement over beam search of 0.2\% and 0.1\% absolute in the single utterance setting. During re-scoring the TLM benefits from interpolation with the first-pass models. Consequently, this is likely due to the TLMs much higher perplexity on this corpora, causing greater reliance on the n-gram models. 

As with AMI we find that beam search enables better utilisation of the context with around a 0.1\% decrease over the single-utterance setting on the test data, compared to fairly insignificant decrease of less than 0.1\% for re-scoring. For both decoding methods we see an increase in WER when using 500 tokens of context. Additionally, there is no benefit from the context on the development set when using the model outputs as history. However, for the GTH evaluations there is around a 0.2\% absolute decrease over single-utterance when using beam search suggesting that the poor performance on this split may be due to misleading errors in the history.

\vspace{-2mm}
\section{Conclusion}
\label{sec:theend}
In this work, we examined the benefit of context from prior utterances during the decoding stage of ASR. Up to 500 tokens were found to be helpful when decoding on AMI with reductions in a similar range to prior work. From Tedlium there are much smaller gains from the context. However, it is difficult to tell to what degree this is a result of our method or the prevalence of useful long-range dependencies in this dataset. 

It was found that decoding directly via beam search allowed for greater use of the context, with increasing improvements compared to re-scoring as more context was made available. As we find better results on AMI where the TLMs show a much lower PPL, it is reasonable to assume that increasing the scale of our models would provide improved results on both datasets. Additionally, due to our findings that the use of context is limited during re-scoring due to the first-pass models, it is likely also limited by the AMs implicit language model. Ideally, cross-utterance information would be incorporated during both encoding and decoding stages, which can be investigated as part of future work.

\bibliographystyle{IEEEtran}
\bibliography{mybib}

\end{document}